\documentclass{article}
\usepackage[T1]{fontenc}
\usepackage{microtype}
\usepackage{graphicx}
\usepackage{booktabs}
\usepackage{hyperref}
\usepackage[preprint]{icml2026}
\usepackage{amsmath,amssymb}
\usepackage{xurl}
\hypersetup{
  pdftitle={PixelJev: Towards General-Purpose Visual Decision Models},
  pdfsubject={A unified native-image decision interface with frozen and adapted open multimodal models},
  pdflang={en-US}
}
\icmltitlerunning{PixelJev: Towards General-Purpose Visual Decision Models}

\begin{document}
\twocolumn[
  \icmltitle{From Text Decisions to Pixels: An Study of Jev-Style Visual Choice Model}
  \begin{icmlauthorlist}
    \icmlauthor{Xunlan Zhou}{nju,msra}
    \icmlauthor{Xianliang Yang}{msra}
    \icmlauthor{Li Zhao}{msra}
  \end{icmlauthorlist}
  \icmlaffiliation{nju}{School of Intelligence Science and Technology,
    Nanjing University, Suzhou, China}
  \icmlaffiliation{msra}{Microsoft Research Asia}
  \icmlcorrespondingauthor{Li Zhao}{lizo@microsoft.com}
  \icmlkeywords{PixelJev, visual decision models, multimodal learning, Jev, transfer, calibration}
  \vskip 0.3in
]
\printAffiliationsAndNotice{%
  \raggedright
  \urlstyle{same}%
  Xunlan Zhou is an intern at Microsoft Research Asia.\par
  Author emails (in author order):
  \nolinkurl{wyattzhouxl@smail.nju.edu.cn},
  \nolinkurl{xianya@microsoft.com},
  \nolinkurl{lizo@microsoft.com}.\par
}

\begin{abstract}
Visual software often needs a decision over supplied alternatives rather
than a generated explanation.
We present PixelJev, a native-image decision interface that maps an image,
a task instruction, and a runtime candidate set to a structured choice
and candidate-conditioned probabilities using small open multimodal models.
Its initial realization unifies recognition and multiple-choice visual
question answering through an existing language-model readout, with
separately evaluated options for frozen inference, language-side adaptation,
and held-out calibration.
Across seven benchmark evaluations, 64-shot source adaptation raises Pets
accuracy from 60.13\% to $92.40\pm0.26$\% across optimization seeds and
transfers to natural resampling, new texture labels, and A-OKVQA without
target fitting, while frozen inference already supports both VQA tasks.
A matched prompt-only follow-up on Pets and ScienceQA attributes the
large Pets gain to adaptation and identifies a narrower output-validity
benefit of candidate readout in adapted VQA.
Specialist DINOv2 probes remain stronger on source recognition, frozen 4B
is stronger than adapted 2B on DTD and ScienceQA, and accuracy gains do
not ensure calibrated target probabilities.
These findings establish a working starting point for general-purpose
visual decision models and identify the remaining requirements:
schema robustness, cross-family transfer, and reliable use of visual evidence.
\end{abstract}

\section{Introduction}
\label{sec:introduction}
A visual model deployed inside software is often asked to decide rather
than describe. The caller may need a category, an answer from a changing
option list, or a choice among proposed objects. A fixed classifier exposes
one learned label vocabulary; a conversational vision-language model (VLM)
exposes generated text. Between these interfaces lies a useful abstraction:
the caller provides the decision space at runtime, and the model returns
a valid identifier with scores conditioned on that space.

We call this abstraction a \emph{visual decision model} and introduce
\textbf{PixelJev}, an implementation with native image input and small open
multimodal backbones. Its interface is
\emph{image + instruction + candidate schema $\rightarrow$ choice +
candidate-conditioned probabilities}.
It follows the program-facing motivation of Jev \citep{typesafe2026} and
open decision interfaces such as SemIf \citep{semif}.
The image enters the backbone's vision branch directly, not through a
captioning intermediary.

The long-term objective is \emph{general-purpose visual decisions}: one
interface and a reusable model that serve changing tasks and candidate
semantics without a newly fitted output head for every label space.
We distinguish three requirements: \emph{interface reuse}, \emph{competence
on held-out task families}, and \emph{reliability under changes to the
visual evidence and decision schema}. This paper establishes interface
reuse across recognition and question answering and measures early transfer;
the broader competence and reliability requirements remain open.
``Towards'' in the title marks that distinction.

The mechanism is deliberately simple. PixelJev uses the existing Qwen3.5
vision-language model \citep{qwen35} and reads probabilities from candidate
tokens at the first assistant position. Frozen inference, language-side
LoRA, and scalar calibration are separate choices rather than prerequisites
bundled into a claimed new architecture.
This separation lets us ask what adaptation actually buys over a working
frozen interface, a larger frozen backbone, and specialist visual encoders.

\begin{figure*}[t]
  \centering
  \includegraphics[width=\textwidth]{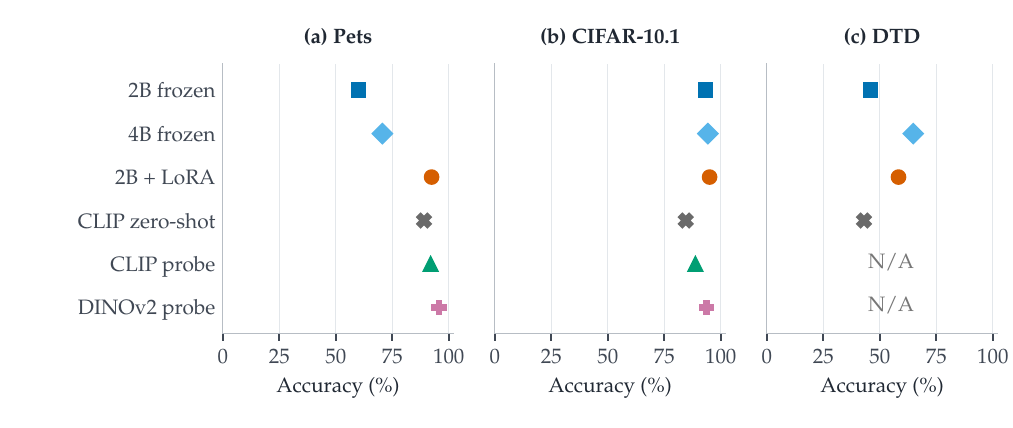}
  \caption{One interface does not imply one best model for every task.
  All points use full test sets. PixelJev's LoRA variant uses 64 fitting
  images per source class and reports the mean of three optimization seeds;
  whiskers show one sample SD, often narrower than the marker.
  Probes use 64-shot source supervision; their heads transfer to CIFAR-10.1,
  but no texture head is fitted for DTD. Missing heads are N/A, not zero.
  Exact values for all five datasets appear in Table~\ref{tab:recognition}.}
  \label{fig:recognition}
\end{figure*}

Related work already explores native visual decisions.
In particular, Visual Jev \citep{yu2026visualjev} combines language-model-head
readout, answer-supervised adaptation, and shared-context execution.
PixelJev addresses a complementary empirical question: how far does
\emph{classification-only, few-shot} adaptation of a small model transfer,
and when is frozen scale or a specialist representation a better choice?
We do not claim priority for the general idea of visual decision interfaces.

Our contributions are threefold.
\textbf{First}, we specify and implement a reusable decision contract with
native pixels, runtime candidate descriptions, stable return identifiers,
and explicit probability semantics.
\textbf{Second}, we evaluate its frozen and adapted realizations across
source recognition, natural resampling, new texture labels, and direct VQA,
with larger-model, specialist, and matched prompt-only alternatives.
\textbf{Third}, we map the limits of reuse: adaptation helps selected
transfers but does not universally replace scale, and improved recognition
does not establish reliable confidence.

\section{Related Work}
\label{sec:related}
\paragraph{Program-facing and visual decisions.}
Jev exposes typed decisions to programs \citep{typesafe2026}, while SemIf
implements open-model candidate-logit readout \citep{semif}.
Early studies examine memory control \citep{jiang2026jevmem}, scientific
choices \citep{deng2026scientific}, probabilistic annotation
\citep{rafe2026calibrated}, and adapted scam-screening decisions
\citep{ren2026openjev}.
Visual Jev \citep{yu2026visualjev} is directly related: it uses a Qwen3-VL
backbone, ordinary answer supervision through the language-model head,
and batched question suffixes sharing an image/context prefix.
It separates adaptation from serving efficiency and tests typed heads as
controls. Our study instead uses Qwen3.5, candidate-conditioned fitting on
recognition labels, specialist probes, and classification-to-VQA transfer.
These differences define scope, not measured superiority over Visual Jev;
we have not run its system under a matched protocol.

\paragraph{Representations and readouts.}
CLIP aligns image and text representations for zero-shot recognition
\citep{radford2021clip}; DINOv2 supplies strong image features
\citep{oquab2024dinov2}. LoRA enables parameter-efficient adaptation
\citep{hu2022lora}. PixelJev assembles established components around a
programmatic contract. Its candidate scoring is closely related to
ordinary multiple-choice VLM inference; a distinct interface does not
establish a distinct learning algorithm or a new visual representation.

\paragraph{Transfer and probability quality.}
Temperature scaling estimates confidence adjustments from held-out labels
\citep{guo2017calibration}.
CIFAR-10.1 tests natural resampling \citep{recht2018cifar}, DTD changes
texture semantics \citep{cimpoi2014dtd}, and A-OKVQA
\citep{schwenk2022aokvqa} and ScienceQA \citep{lu2022scienceqa} change
questions and answer choices per example. We use these settings to separate
interface reuse, source adaptation, and target probability quality.
None establishes exclusion from unknown foundation-model pretraining.
\section{PixelJev: A Reusable Visual Decision Interface}
\label{sec:method}
\subsection{The Decision Contract}
A request comprises an RGB image $x$, an instruction or question $q$, and
an ordered candidate schema $C=((u_1,c_1),\ldots,(u_K,c_K))$.
Here $u_i$ is a stable programmatic identifier and $c_i$ is its textual
description. A decision model implements
\begin{equation}
  F_\theta(x,q,C)=(u_{\hat{\imath}},\mathbf{p},m),
  \qquad \hat{\imath}=\arg\max_i p_i,
  \label{eq:contract}
\end{equation}
where $\mathbf{p}$ is a distribution over the supplied candidates and $m$
records model, schema, and calibration provenance.
Identifiers are returned by the program, not generated as free text.

The same contract covers different workloads without equating their
difficulty. Recognition keeps $q$ and the class descriptions fixed across
images. Multiple-choice VQA supplies a new $q$ and new descriptions for
each question. New-class evaluation supplies descriptions outside adapter
supervision without fitting a new classifier head.
Object and action candidate selection are possible future uses of the
contract, but are not evaluated capabilities of this study.

The implementation validates the schema before scoring. Legal output
membership follows from the programmatic readout; correctness must be
measured separately. Likewise, $\mathbf{p}$ is conditional on $C$, not a
probability that the selected answer is correct among all conceivable
alternatives. There is no learned reject option in the current system.

\subsection{An Initial Realization with Open Multimodal Models}
The complete Qwen3.5 image processor and vision branch consume $x$.
The user message contains $q$ and the ordered descriptions; filenames,
sample identifiers, correct labels, and explanatory answers are excluded.
There is no captioning intermediary.
Each candidate maps to a case-sensitive alphabetic slot $s_i$.
The tokenizer checks single-token reversibility, distinct token IDs, and
preservation of the completed chat-template boundary.
The current slot inventory supports $2\le K\le52$; this is an implementation
limit, not a definition of visual decision modeling.

With extended thinking disabled, let $z$ be vocabulary logits at the first
assistant position. PixelJev reads
\begin{equation}
  p_i(T\mid x,q,C)=
  \frac{\exp(z_{s_i}/T)}{\sum_{j=1}^{K}\exp(z_{s_j}/T)}.
  \label{eq:choice}
\end{equation}
Raw inference sets $T=1$.
The total vocabulary probability mass on candidate tokens is logged
separately from this renormalized distribution.
No response string or JSON document is generated for subsequent parsing.

\paragraph{Relationship to ordinary VLM inference.}
Equation~\ref{eq:choice} reuses an existing multiple-choice readout, rather
than introducing a new architecture.
PixelJev makes its input/output contract and candidate-conditional
semantics explicit and evaluates how a reusable model behaves across
workloads. Direct logits match equivalent constrained, deterministic
one-token generation on the diagnostic inputs.
We therefore do not attribute an intrinsic speedup to omitting that token
or claim a task-accuracy advantage over all conventional VLM prompting.
The full-set prompt-only follow-up crosses this readout with the same
checkpoints under a common format cue and separately measures warm
inference time (Table~\ref{tab:prompt-baseline};
Appendix~\ref{app:prompt-baseline}).

\subsection{Three Separately Evaluated Operating Regimes}
\paragraph{Frozen inference.}
The complete pretrained checkpoint and the validated slot mapping are
sufficient to instantiate Equation~\ref{eq:contract}. No benchmark fitting
labels or extra decision head are required. Both 2B and 4B backbones use
the same principal interface and pixel budget.

\paragraph{Few-shot adaptation.}
We optionally minimize candidate-conditioned cross-entropy,
\begin{equation}
  \mathcal{L}(x,y,q,C)=-\log p_y(1\mid x,q,C),
\end{equation}
randomly permuting candidates and remapping the correct slot during fitting.
One LoRA adapter is shared across the three source recognition tasks.
Only explicit language-attention projection allowlists are updated;
the original vision, language, embedding, and output parameters remain
frozen. We use rank 16, scaling 32, dropout 0.05, and three AdamW epochs
at learning rate $10^{-4}$.
Pets development accuracy selects the checkpoint, with raw NLL breaking
ties, and that checkpoint is reused across all evaluation tasks.
Appendix~\ref{app:protocol} records the full supervision ledger.

\paragraph{Held-out calibration.}
A positive scalar temperature is fitted by minimizing NLL on a dedicated
calibration split. This consumes labels even though it does not update
the backbone or change the argmax.
Raw and calibrated evaluations remain separate.
Only the source CIFAR-10 temperature transfers to CIFAR-10.1; no DTD or
VQA temperature is fitted.
Frozen inference, adapter fitting, and calibration are thus distinct costs
and scientific interventions, not interchangeable meanings of ``training.''

\subsection{Interface Diagnostics}
Candidate reversal, an alternate instruction, and actual different-class
image substitution provide finite interface diagnostics.
Image-substitution controls and one-token readout comparisons are reported in
Appendix~\ref{app:checks}. They test finite interface behavior, not
arbitrary schema invariance or visual evidence dependence on every task.
\section{Evaluating Reuse, Adaptation, and Transfer}
\label{sec:protocol}
The evaluation follows the intended progression of a reusable decision
model: instantiate the interface without fitting, adapt on a restricted
source task family, and test reuse beyond those labels.
Source recognition, natural shift, new texture semantics, and per-question
VQA choices answer different questions; we do not collapse them into one
``generality'' score.
\begin{table*}[t]
  \caption{Full-set recognition accuracy (\%). LoRA reports the mean $\pm$
  sample SD over seeds 7, 8, and 9; the other rows are single fixed-model
  evaluations. Both probes use 64 fitting examples per source class.
  CIFAR-10.1 reuses source heads; DTD receives no support labels.
  The three left columns are source tasks, and the two right columns are
  transfer tasks. No target fitting or test-based model selection is used.}
  \label{tab:recognition}
  \centering
\begin{tabular}{@{}lccccc@{}}
\toprule
Method & Pets & CIFAR-10 & EuroSAT & CIFAR-10.1 & DTD \\
\midrule
2B frozen & 60.13 & 96.04 & 49.63 & 93.20 & 45.85 \\
4B frozen & 70.65 & 96.92 & 56.81 & 94.30 & 64.79 \\
2B + LoRA (64-shot) & $92.40\pm0.26$ & $97.04\pm0.11$ & $88.31\pm1.57$ & $95.05\pm0.13$ & $58.28\pm0.98$ \\
CLIP zero-shot & 89.07 & 90.13 & 47.67 & 84.50 & 42.98 \\
CLIP probe & 91.99 & 93.38 & 91.39 & 88.75 & --- \\
DINOv2 probe & 95.67 & 97.62 & 92.13 & 93.75 & --- \\
\bottomrule
\end{tabular}

\end{table*}

\subsection{Source Tasks and Supervision}
We use CIFAR-10 object recognition \citep{krizhevsky2009}, all 37 Oxford-IIIT
Pets breeds \citep{parkhi2012pets}, and EuroSAT land-cover classification
\citep{helber2019eurosat}. We retain each pinned mirror's full official
test partition. EuroSAT uses a fixed mirrored partition, without a
geographical-generalization claim.

Exact identity is computed from decoded RGB pixels and image dimensions.
A fixed perceptual-hash policy groups near-similar images across complete
training and test pools before selecting fitting examples.
Training groups touching test data are excluded from adaptation, and
fitting, development, and calibration groups are disjoint.
All test images remain in the denominator; correlated images are handled
through grouped uncertainty rather than removal.
This policy does not prove that all leakage or pretraining overlap is absent.

The two budgets contain 16 or 64 fitting images per class, totaling 912 or
3,648 images across tasks. The smaller set is a strict subset of the larger.
There are also 684 development and 684 calibration images, unchanged across
budgets. Thus ``64-shot'' counts fitting labels, not all labels used by the
study. The 64-shot condition is repeated with optimization seeds 7, 8, and 9
on the same split; none of the reported models is a seed ensemble.

\subsection{Alternatives to Adapting the Interface}
We evaluate full Qwen3.5-2B and Qwen3.5-4B checkpoints under the same
principal user-message interface and native-image pixel budget
(65,536 minimum; 262,144 maximum).
We also evaluate frozen CLIP ViT-B/16 zero-shot scores and logistic-regression
heads on frozen, normalized CLIP and DINOv2 ViT-B features.
Each probe uses the same per-task fitting images, with regularization
selected on development data. Each backbone retains its native processor.

The shared Qwen adapter sees labels from all three tasks; each specialist
head sees labels from its own task only. Per-task fitting shots match, but
total supervision, trainable parameter count, and compute do not.
Raw frozen Qwen and zero-shot CLIP consume no benchmark fitting labels;
temperature-scaled variants additionally consume calibration labels.
Our comparisons test practical alternatives to small-model adaptation.
They are not compute-matched architectural ablations.
The matched prompted follow-up below isolates a fixed checkpoint's
inference path; complete-candidate likelihood and independently optimized
reasoning prompts remain outside scope.

\subsection{Matched Prompt-Only Follow-up}
We cross frozen 2B / the existing seed-7 adapter with ordinary greedy
generation / direct candidate readout on the complete Pets and ScienceQA
selections, representing source recognition and VQA transfer.
Both paths receive identical images, options, and a fixed system
format instruction. Ordinary generation has no candidate-token mask.
The system cue was selected by output-format compliance, not accuracy,
on eight preselected Pets development images.
Direct inference is rerun under that cue, rather than compared against
the original user-only prompt.
Invalid generated answers count as errors. A fixed answer-line parser
permits preceding explanation, with a 128-token cap.
Warm batch-one timing interleaves direct, ordinary-generation, and
constrained-one-token requests on a locked subset.
Appendix~\ref{app:prompt-baseline} specifies parsing, format selection,
timing boundaries, and the limits of attribution.
This targeted follow-up does not establish readout effects on the other
five benchmarks.

\subsection{Transfer Without Target Fitting}
\paragraph{Natural resampling.}
CIFAR-10.1 v6 contains 2,000 images in the existing ten categories.
We reuse the original task instruction, candidate descriptions, source
probe heads, and source temperatures. This is a relatively mild natural
resampling shift, not an adversarial or corruption stress test.

\paragraph{New task and label space.}
DTD uses all 1,880 images in the official partition-1 test set and all 47
texture labels. No DTD fitting, development, or calibration labels are used.
CLIP uses a fixed texture prompt. Ordinary DINOv2 has no text mapping for
the new classes, so its unsupported head-free result is N/A.
DTD expands beyond the 37-slot adaptation vocabulary: ten slots were never
positive targets during fitting. The evaluation therefore changes visual
content, semantics, and candidate cardinality simultaneously.
``Out of distribution'' refers to adapter supervision, not unknown
foundation-model pretraining.

\paragraph{Question-conditioned choices.}
A-OKVQA uses all 1,145 public validation questions; ScienceQA uses every
image-containing test question, 2,017 of 4,241.
We compare frozen 2B, frozen 4B, and the already selected classification
adapter with predesignated seed 7. There is no VQA fitting, demonstration
selection, prompt search, temperature estimation, or checkpoint selection.
Inputs contain the image, question, ordered choices, and ScienceQA's
hint/context. Rationales, direct answers, lectures, and solutions are
excluded. A-OKVQA follows official answer-text scoring, including repeated
options; ScienceQA follows official answer-index scoring.

A separate frozen CLIP follow-up scores the same choices by image/text
similarity using a fixed \texttt{Question--Context--Answer} template.
Within its 77-token window, we reserve every complete candidate answer,
then allocate the same question/context prefix to all choices.
In these evaluations no question or answer is truncated; ScienceQA context
is shortened for 424 questions.
This is contrastive matching, not a trained VQA head, and CLIP's native
image crop is not resolution-matched to Qwen.
Plain DINOv2 cannot score arbitrary answer text without an additional
text-alignment mechanism.

\subsection{Selection and Uncertainty}
The 64-shot budget and checkpoint selection use development endpoints.
The scale/OOD and VQA protocols are follow-ups fixed before their respective
predictions, not a claim of project-wide prospective preregistration.
No target labels guide adaptation, prompts, or selection.

Accuracy is primary; we also retain macro accuracy, NLL, multiclass Brier
score, 15-bin expected calibration error (ECE), and selective risk.
Reported LoRA spreads are sample standard deviations across optimization
seeds, not uncertainty across training splits.
Method differences use 2,000 paired bootstrap resamples of image-similarity
groups, preserving all questions attached to sampled groups.
Intervals are descriptive 95\% intervals without multiplicity correction.
Comparisons preserve complete evaluation sets and align images, labels,
and image-similarity groups across methods.
\section{What Transfers Through a Shared Interface?}
\label{sec:results}
We evaluate the same decision mechanism with different backbones and an
optional source-trained adapter. The results distinguish a usable frozen
interface from the stronger claim that one adaptation recipe improves
every workload.

\subsection{Source Fitting Improves the Small Model, Not Every Model Ranking}
Table~\ref{tab:recognition} shows the largest adaptation gains on Pets and
EuroSAT. Full-class Pets accuracy rises from 60.13\% for frozen 2B to
$92.40\pm0.26$\% after adaptation. EuroSAT rises from 49.63\% to
$88.31\pm1.57$\%. CIFAR-10 has less headroom: the corresponding change is
96.04\% to $97.04\pm0.11$\%.

The specialist comparison puts these adaptation gains in context.
DINOv2's 64-shot probe has the highest point estimate on all three source
tasks: 95.67\%, 97.62\%, and 92.13\%, respectively.
On Pets, adapted-minus-CLIP-probe paired intervals include zero for every
seed, whereas the adapted models remain below DINOv2.
Interface flexibility and fixed-label accuracy are therefore separate
selection criteria. Reusing a language-conditioned interface need not be
the best choice when the application only requires one fixed classifier.

Increasing frozen scale improves source-task point estimates to 70.65\%,
96.92\%, and 56.81\%. Scale alone does not close the Pets or EuroSAT
adaptation gap, but frozen 4B is close to the adapted small model on
CIFAR-10. These are results of the specified letter-choice interface,
not estimates of the best unconstrained prompting strategy for each model.

\subsection{Source Adaptation Extends Beyond Its Label Vocabulary}
On CIFAR-10.1, adapted 2B reaches $95.05\pm0.13$\%, improving over frozen
2B's 93.20\% for every seed with paired intervals above zero.
The source-task specialist ranking changes: DINOv2 reaches 93.75\%.
However, seed 7's advantage over DINOv2 is only 1.25 percentage points (pp),
with an unadjusted interval of $[0.05,2.50]$.
The lower endpoint is close to zero, and these exploratory comparisons do
not support broad superiority claims.
Against frozen 4B, two of the three adapted-seed intervals include zero;
we do not select the remaining seed to claim a robust win.

On DTD, adapted accuracy is $58.28\pm0.98$\%, above frozen 2B at 45.85\%
but below frozen 4B at 64.79\%. Every seed improves over the smaller
frozen base and remains below the larger frozen model, with the respective
paired intervals excluding zero.
Adaptation therefore transfers beyond its supervised label vocabulary,
but does not substitute uniformly for frozen model scale.
This is not evidence of adaptation-induced visual forgetting:
accuracy improves relative to the same frozen backbone, and multiple
task factors change together.

No target images are flagged by the specified adaptation-overlap check on
CIFAR-10.1 or DTD. Consequently, the declared overlap-exclusion sensitivity
subsets coincide with the full sets. This is a result of one particular
check, not proof of unrestricted data independence.

\begin{figure*}[!t]
  \centering
  \includegraphics[width=\textwidth]{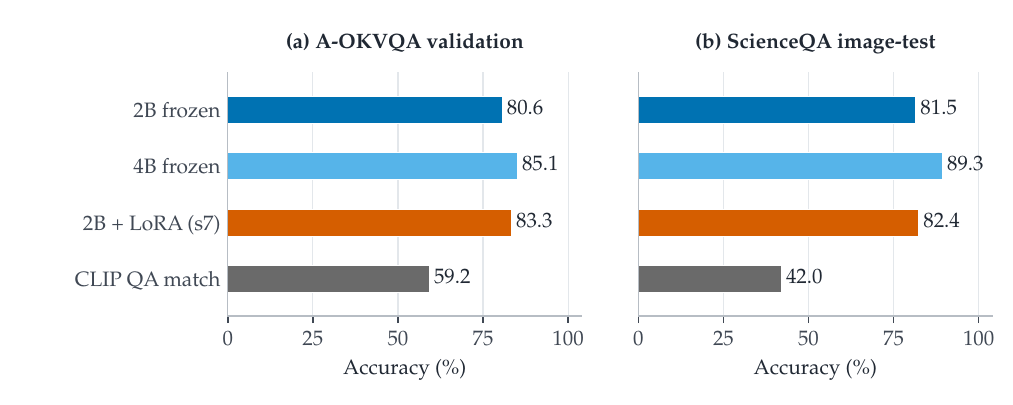}
  \caption{Direct multiple-choice VQA without VQA training, calibration,
  or prompt selection. CLIP uses fixed question--context--answer matching;
  LoRA is the existing classification adapter with predesignated seed 7.
  Bar lengths start at zero and show point estimates, not seed averages.
  Paired uncertainty is shown in Figure~\ref{fig:paired}.
  A-OKVQA is public validation, not the hidden-label test leaderboard.
  Exact two-decimal results appear in Appendix Table~\ref{tab:vqa}.}
  \label{fig:vqa}
\end{figure*}

\subsection{Per-Question Schemas Work Without VQA Fitting}
Figure~\ref{fig:vqa} tests reuse when both the question and candidate
semantics change per example. No new VQA head or adapter is fitted.
Frozen 2B achieves 80.61\% on A-OKVQA and 81.51\% on ScienceQA;
frozen 4B achieves 85.07\% and 89.34\%.
The classification adapter reaches 83.32\% and 82.35\%, respectively.

Figure~\ref{fig:paired} distinguishes point-estimate gains from resolved
paired differences. On A-OKVQA, adaptation improves over frozen 2B by
2.71 pp, with interval $[1.13,4.30]$.
Its difference from frozen 4B is $-1.75$ pp with interval $[-3.68,0.26]$,
so the larger model's point-estimate lead is not a resolved paired difference.
On ScienceQA, the adapter's change from frozen 2B is $+0.84$ pp,
with interval $[-2.14,3.33]$; this does not establish an adaptation benefit.
The adapted-minus-4B difference is $-6.99$ pp,
with interval $[-10.73,-4.08]$.

Frozen CLIP question-answer matching reaches 59.21\% on A-OKVQA and
41.99\% on ScienceQA. Every evaluated Qwen variant exceeds it on both
datasets with paired intervals above zero.
CLIP can rank textual choices, but this fixed contrastive formulation is
weaker than the evaluated multimodal language models on these workloads.
It is not an upper bound on trained CLIP-based VQA systems.
Different image preprocessing and CLIP's shortened ScienceQA contexts also
prevent attributing the entire gap to reasoning ability.
The interface extends successfully to this workload, but source fitting
has not acquired uniformly stronger question-answering competence.
These comparisons use one fixed adapter seed; their intervals do not
cover adapter-training variability.

\begin{table*}[t]
  \caption{Matched checkpoint $\times$ readout follow-up: full-set accuracy
  (\%). Gen.\ is unconstrained greedy generation with a fixed parser;
  Read.\ is independently executed candidate-logit inference.
  Frozen Gen.\ is the prompt-only baseline; LoRA Gen.\ uses the existing
  classification adapter (seed 7), not new training.
  All four cells use the same system format cue and are separate from the
  original user-only-prompt results.
  The follow-up is limited to Pets and ScienceQA.
  Invalid F/L counts are frozen/adapted generation failures, included as
  incorrect in the full denominators.}
  \label{tab:prompt-baseline}
  \centering
\begin{tabular}{@{}lrrrrrl@{}}
\toprule
Dataset & $n$ & \multicolumn{2}{c}{Frozen 2B} & \multicolumn{2}{c}{2B + LoRA} & Invalid F / L \\ & & Gen. & Read. & Gen. & Read. & \\
\midrule
Pets & 3,669 & 56.36 & 56.36 & 91.63 & 91.63 & 0 / 0 \\
ScienceQA & 2,017 & 77.94 & 77.94 & 78.93 & 80.81 & 0 / 55 \\
\bottomrule
\end{tabular}

\end{table*}

\subsection{Prompt-Only Controls Separate Adaptation from Output Validity}
Table~\ref{tab:prompt-baseline} shows that Pets' gain follows the adapter,
not a different answer path: both paths rise from 56.36\% to 91.63\%,
or $+35.27$ pp ($[33.47,37.03]$), with identical decisions at fixed weights.
On ScienceQA, frozen paths also agree at 77.94\%.
The adapter reaches 78.93\% with generation and 80.81\% with direct readout.
Its generated improvement is unresolved ($+0.99$ pp,
$[-1.18,2.87]$), whereas direct-minus-generated accuracy is $+1.88$ pp
($[1.13,2.86]$).

This latter difference is entirely an output-validity effect in these
records: the adapted generator emits 52 out-of-range letters and three
other invalid formats; valid generated decisions agree with direct
readout. Direct readout answers 38 of those 55 invalid cases correctly.
This is evidence for enforcing the supplied decision space, not improved
visual reasoning from a new head. Alternative parsing or full-vocabulary
answer SFT remains untested.

Ordinary generation takes 1.32--1.44 times the mean warm direct-request
time across these four cells; it generates approximately two tokens per
response. The equivalent constrained-one-token path takes only
1.03--1.04 times direct time.
Thus the measured saving primarily avoids response decoding/termination
and framework work, not an inherently cheaper decision computation.
Appendix Table~\ref{tab:prompt-timing} gives absolute times and boundaries.
The changed format cue also changes accuracy relative to the original
user-only interface; the two tables must not be mixed to attribute gains.

\subsection{Probability-Valued Outputs Still Need Reliability Evidence}
Calibration often lowers held-out NLL, but not every metric improves.
The Pets DINOv2 probe worsens slightly in both NLL and ECE; frozen CIFAR-10
and adapted EuroSAT also worsen in ECE despite improved NLL.
Appendix Table~\ref{tab:calibration} preserves these negative cases;
Figure~\ref{fig:calibration} shows the calibrated Pets reliability and
risk-coverage curves.

On DTD, adaptation lowers mean raw NLL from the frozen base's 2.6985 to
1.9253, while mean raw ECE changes from 0.1905 to 0.2040.
Thus target accuracy gains do not imply automatically calibrated target
probabilities. We do not fit a target temperature to repair this result.
NLL comparisons here are within a dataset and candidate set, not cross-task
rankings across different label cardinalities.

\paragraph{Adaptation is an additional cost, not an interface prerequisite.}
The fitting-budget comparison shows the largest dependence on satellite
imagery (Appendix Table~\ref{tab:budget}).
The adapter has 7,382,016 trainable parameters; 64-shot summed optimizer-step
intervals span 3,220.25--3,373.94 seconds, excluding model loading,
development evaluation, saving, and parameter-integrity checks.
Appendix~\ref{app:cost} separates this training cost from inference latency.
\section{Discussion: What Does ``General-Purpose'' Require?}
\label{sec:limitations}
\paragraph{Reuse is demonstrated; general competence is a goal.}
PixelJev uses the same contract and candidate-token readout for fixed
recognition vocabularies and changing VQA options. Source-trained adapters
are reused without target fitting. These are concrete forms of reuse.
They do not establish competence on arbitrary visual tasks, nor demonstrate
detection, segmentation, action selection, or open-world rejection.
The observed specialist and frozen-scale advantages locate the present
implementation on a broader design space rather than invalidate its
interface.

\paragraph{The boundary to existing methods matters.}
Candidate-logit readout, LoRA, and temperature scaling are established;
Visual Jev already combines native images, adaptation, and shared-context
decision execution \citep{yu2026visualjev}.
Our evidence concerns classification-to-decision transfer, frozen model
scale, and specialist alternatives, not a first visual decision architecture
or a reproduction of proprietary RLCD.
No matched experiment against Visual Jev is reported.
The frozen visual tower also means that accuracy gains do not by themselves
demonstrate improved visual representations.
The prompt-only factorial separates the fixed adapter from its inference
path, not candidate-conditioned training from ordinary answer SFT.

\paragraph{Current transfer evidence has specific limits.}
One adaptation split, three optimization seeds, and one VQA adapter seed
do not cover training-data or model-family variation.
Foundation-model pretraining overlap is unknown.
DTD changes images, semantics, candidate count, and supervised slot coverage
together; its key attribute can omit other valid descriptions.
EuroSAT may retain spatial correlations beyond image-hash grouping.
VQA questions may be answerable from text or world knowledge alone,
so image-containing accuracy is not proof of visual evidence use.
Native processors and total training supervision differ across baselines.
The fixed nonthinking interface is not each backbone's best possible
unrestricted reasoning strategy.

\paragraph{Candidate confidence is not a trust decision.}
The distribution excludes unlisted alternatives, and changing the candidate
set changes the conditioning event. Scores need not be comparable across
different schemas. Low ECE alone is insufficient, and detecting missing
visual evidence is not equivalent to predicting answer error.
The current diagnostics establish finite readout behavior, not calibrated
deployment risk.

\section{A Research Agenda for Visual Decision Models}
\label{sec:agenda}
The following directions are proposed work, not additional results.
\textbf{Schema robustness} should be tested by semantic-ID-aligned
permutations, paraphrases, and controlled distractor changes, separating
slot/token artifacts from genuine changes in task difficulty.
\textbf{Cross-family transfer} requires training on multiple decision
families while withholding complete families and their image groups,
with matched ordinary VLM and specialist baselines.
\textbf{Evidence-aware decisions} require relevant-region interventions,
matched irrelevant-region controls, and separate targets for answer
correctness versus the need to acquire a better observation.
\textbf{Deployment value} must be measured through end-to-end risk,
latency, and memory under a concrete workload beyond the present warm,
single-request measurements; shared-context execution
should be compared with already-batched and cached baselines rather than
claimed from one-token readout alone.
Appendix~\ref{app:agenda} gives falsifiable milestones and a staged order.

\section{Conclusion}
PixelJev provides a working native-image decision interface whose candidate
schema is supplied at runtime rather than fixed by a task-specific output
head. Small frozen multimodal models already instantiate this interface;
few-shot language-side adaptation improves source recognition and selected
transfers without VQA-specific fitting.
Specialist encoders, larger frozen models, and held-out calibration reveal
where this initial realization remains limited.
The path toward general-purpose visual decision models is therefore not
simply to improve one classification score: it is to preserve useful
decisions as task semantics, candidate schemas, and visual evidence change.

\section*{Impact Statement}
This work examines bounded visual decisions on existing research benchmarks.
A syntactically valid choice can still be incorrect, and confidence normalized
over supplied alternatives does not account for an omitted correct answer.
The system should not be treated as a validated decision-maker for medical,
legal, safety-critical, or surveillance applications.
Pretraining overlap, benchmark bias, and performance across demographic groups
are not resolved by these experiments. Model and dataset use remains subject
to the original licenses and access conditions.

\bibliography{references}
\bibliographystyle{icml2026}

\clearpage
\appendix
\onecolumn
\section{Supervision, Selection, and Implementation}
\label{app:protocol}
\subsection{Sample Ledger}
Table~\ref{tab:ledger} distinguishes fitting labels from development and
calibration labels. Selection uses split seed 20260923 throughout.
The 16-shot fitting set is a strict subset of the 64-shot set.
Development, calibration, and test selections are identical across budgets.
The shared adapter sees 912 or 3,648 fitting images in total, plus development
labels for checkpoint selection. The additional 684 calibration labels
are used only for probability scaling, not for adapter updates.

\begin{table}[h]
  \caption{Source supervision and complete transfer denominators.
  Dashes indicate that no target fitting, development, or calibration split
  is used. Pets includes every breed.}
  \label{tab:ledger}
  \centering
  \begin{tabular}{@{}lrrrrr@{}}
    \toprule
    Dataset & Classes & Fit: 16-shot / 64-shot & Development & Calibration & Test \\
    \midrule
    Oxford-IIIT Pets & 37 & 592 / 2,368 & 444 & 444 & 3,669 \\
    CIFAR-10 & 10 & 160 / 640 & 120 & 120 & 10,000 \\
    EuroSAT & 10 & 160 / 640 & 120 & 120 & 5,400 \\
    CIFAR-10.1 v6 & 10 & --- & --- & --- & 2,000 \\
    DTD partition 1 & 47 & --- & --- & --- & 1,880 \\
    \bottomrule
  \end{tabular}
\end{table}

We hash decoded RGB pixels together with image dimensions for exact identity.
The perceptual hash converts the image to $32\times32$ grayscale, applies a
discrete cosine transform, and thresholds the upper-left $8\times8$ coefficients
after excluding the DC component. Connected groups use Hamming distance at
most four. This grouping is computed over the full selected mirror's
official training and test pools before adaptation sampling.
Training groups touching test are excluded; within remaining training groups
we retain a representative and exclude conflicting-label groups.
The official test denominator is never reduced to make results appear cleaner.
Image-similarity grouping is conservative and does not establish that every
pair in a connected group is a true duplicate.

\subsection{Optimization and Scoring Settings}
\begin{table}[h]
  \caption{Fixed optimization and scoring settings.}
  \label{tab:hyperparameters}
  \centering
  \begin{tabular}{@{}ll@{}}
    \toprule
    Setting & Value \\
    \midrule
    LoRA rank / scaling / dropout & 16 / 32 / 0.05 \\
    Optimizer / learning rate / weight decay & AdamW / $10^{-4}$ / 0.01 \\
    Microbatch / gradient accumulation / gradient norm limit & 1 / 16 / 1.0 \\
    Precision / epochs & BF16 / 3 \\
    Gradient checkpointing & Enabled \\
    Fitting visits & Every fitting image once per epoch \\
    Optimization seeds / split seed & 7, 8, 9 / 20260923 \\
    Trainable / complete parameters, including adapter & 7,382,016 / 2,220,623,680 \\
    Principal minimum / maximum pixels & 65,536 / 262,144 \\
    Frozen high-resolution control maximum pixels & 1,048,576 \\
    Probe regularization grid & $C\in\{0.01,0.1,1,10,100\}$ \\
    Probe solver / maximum iterations & LBFGS / 3,000; nonconvergence is an error \\
    Temperature bounds / ECE bins & $[0.05,20]$ / 15 equal-width bins \\
    Paired bootstrap repetitions / seed & 2,000 / 20260923 \\
    \bottomrule
  \end{tabular}
\end{table}

The explicit language-side projection allowlist contains
\texttt{q\_proj}, \texttt{k\_proj}, \texttt{v\_proj}, \texttt{o\_proj},
\texttt{in\_proj\_qkv}, \texttt{in\_proj\_z}, \texttt{in\_proj\_b},
\texttt{in\_proj\_a}, and \texttt{out\_proj}.
No original backbone tensor is trainable.
All three selected 64-shot positive-training checkpoints are from epoch 3.
The fixed visual pixel budget is a processor bound rather than a claim
that every image has identical dimensions.
The frozen higher-resolution control increases only the upper bound and
does not materially improve Pets accuracy.

CLIP/DINO feature vectors are L2-normalized; DINO uses the final CLS
representation. Each probe fits its own task labels, unlike the shared
multitask Qwen adapter. Native backbone preprocessing is retained, so
aligned examples do not imply equal-resolution or equal-crop inputs.

\section{Additional Probability and Budget Results}
\label{app:probability}
\begin{table}[h]
  \caption{Held-out probability quality. Subscript $T$ indicates separate
  source-label temperature calibration; all LoRA rows use seed 7.
  NLL and ECE are unitless and lower is better within a dataset.
  Calibration does not change accuracy.
  The Pets DINOv2 row and the ECE increases on CIFAR-10/EuroSAT show
  that calibration is not uniformly beneficial across held-out metrics.}
  \label{tab:calibration}
  \centering
\begin{tabular}{@{}lrrrr@{}}
\toprule
Task / method & NLL & NLL$_T$ & ECE & ECE$_T$ \\
\midrule
Pets / 2B frozen & 1.8019 & 1.5856 & 0.1373 & 0.0684 \\
Pets / LoRA & 0.2584 & 0.2502 & 0.0290 & 0.0143 \\
Pets / DINOv2 probe & 0.1451 & 0.1461 & 0.0055 & 0.0101 \\
CIFAR-10 / 2B frozen & 0.1664 & 0.1521 & 0.0148 & 0.0165 \\
CIFAR-10 / LoRA & 0.1090 & 0.1047 & 0.0095 & 0.0068 \\
EuroSAT / LoRA & 0.3323 & 0.3271 & 0.0165 & 0.0222 \\
\bottomrule
\end{tabular}

\end{table}

\begin{figure}[h]
  \centering
  \includegraphics[width=\textwidth]{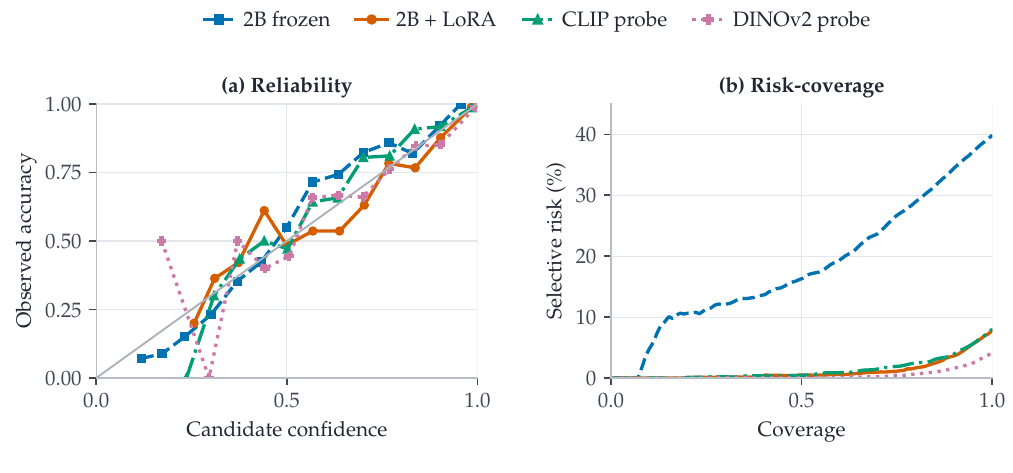}
  \caption{Pets test reliability and selective risk after source calibration.
  Reliability points are occupied equal-width confidence bins; connecting
  segments are guides, not smoothed fits or confidence intervals.
  All methods use their own independently fitted source temperature.
  LoRA uses seed 7. Risk is the error rate among predictions retained in
  descending confidence order; curves show the measured evaluation points.
  Better uncertainty behavior is an empirical property, not a consequence
  of returning a probability-valued schema.}
  \label{fig:calibration}
\end{figure}

\begin{table}[h]
  \caption{Single-seed fitting-budget comparison, accuracy (\%).
  Development, calibration, and test sets are unchanged; this is not a
  multi-seed data-scaling experiment.}
  \label{tab:budget}
  \centering
\begin{tabular}{@{}lrrr@{}}
\toprule
Fitting budget & Pets & CIFAR-10 & EuroSAT \\
\midrule
16-shot / seed 7 & 89.89 & 96.84 & 67.35 \\
64-shot / seed 7 & 92.20 & 97.00 & 89.67 \\
\bottomrule
\end{tabular}

\end{table}

\subsection{Measured Adaptation Cost}
\label{app:cost}
\begin{table}[h]
  \caption{Logged optimizer-step intervals on a shared NVIDIA GB10 system.
  Sums are not total run wall time.}
  \label{tab:cost}
  \centering
\begin{tabular}{@{}lrrr@{}}
\toprule
Run & Steps & Sum (s) & Median (s) \\
\midrule
16-shot / seed 7 & 171 & 927.51 & 5.40 \\
64-shot / seed 7 & 684 & 3373.94 & 4.87 \\
64-shot / seed 8 & 684 & 3221.01 & 4.72 \\
64-shot / seed 9 & 684 & 3220.25 & 4.70 \\
\bottomrule
\end{tabular}

\end{table}

Intervals include accumulated microbatches and image/input preparation.
They exclude loading, epoch-end development evaluation, checkpoint saving,
and frozen-parameter integrity checks. They are neither exclusive GPU kernel
times nor a cost accounting of the whole investigation.
The pinned runtime is NGC PyTorch 25.11 with Python 3.12.3,
PyTorch \texttt{2.10.0a0+b558c986e8.nv25.11}, and CUDA 13.
Experiments use Transformers 5.17.0, PEFT 0.18.1, Datasets 5.0.1,
and scikit-learn 1.7.2.
The runtime container is limited to eight CPUs and 48 GB of memory;
the underlying machine is shared.

\section{Direct Multiple-Choice VQA}
\label{app:vqa}
\begin{table}[h]
  \caption{Complete question selections. ScienceQA image inclusion depends
  only on source image presence, not model outcomes. The two sources have
  no images flagged by the prescribed adaptation-overlap check.}
  \label{tab:vqaledger}
  \centering
  \begin{tabular}{@{}lrrl@{}}
    \toprule
    Evaluation & Questions & Distinct decoded images & Choices per question \\
    \midrule
    A-OKVQA public validation & 1,145 & 1,122 & 4 throughout \\
    ScienceQA image-test & 2,017 & 1,799 & 2: 677; 3: 532; 4: 770; 5: 38 \\
    \bottomrule
  \end{tabular}
\end{table}

\begin{table}[h]
  \caption{Exact direct VQA accuracy (\%). There is no VQA-specific fitting
  or calibration. The adapter is predesignated classification seed 7.}
  \label{tab:vqa}
  \centering
\begin{tabular}{@{}lrr@{}}
\toprule
Method & A-OKVQA & ScienceQA \\
\midrule
2B frozen & 80.61 & 81.51 \\
4B frozen & 85.07 & 89.34 \\
2B + classification LoRA (seed 7) & 83.32 & 82.35 \\
CLIP frozen QA matching & 59.21 & 41.99 \\
\bottomrule
\end{tabular}

\end{table}

\paragraph{Source and scoring details.}
A-OKVQA uses the validation split of \texttt{HuggingFaceM4/A-OKVQA}.
ScienceQA uses the image-containing test examples from
\texttt{derek-thomas/ScienceQA}.
Images are decoded to RGB and aligned by decoded-pixel identity.

A-OKVQA contains seven questions with repeated option text, including three
whose correct text occurs in multiple slots, and one empty distractor.
All questions and slots are retained. Official multiple-choice scoring
compares answer text, so every slot containing the correct string is accepted.
An empty option is rendered as \texttt{[empty option]} to satisfy the
nonempty-description interface, while its original string and position are
retained for scoring. ScienceQA scores the official answer index.
These are benchmark properties, not model errors or exclusion criteria.

\paragraph{CLIP text budget.}
The frozen contrastive baseline builds
\texttt{Question: ...}, optional \texttt{Context: ...}, and \texttt{Answer: ...}
segments. The 77-token native window includes start/end tokens.
We reserve enough space for the longest complete answer and retain a common
question/context prefix for every option. Shorter answers do not receive
additional context. Candidate answers are never truncated.
All questions and answers fit in both evaluations.
No A-OKVQA text is shortened; 424 ScienceQA contexts are shortened.
The template and packing rule are fixed before predictions.
This baseline has no trained multimodal fusion head and is not intended
as an upper bound on CLIP-based VQA.

The ordinary DINOv2 checkpoint has no aligned text encoder, so it cannot
perform the same arbitrary text-choice matching directly.
Text-aligned variants such as dino.txt introduce additional components
and constitute different checkpoints; they are not evaluated here.
ScienceQA's image-similarity bootstrap uses 1,016 groups, fewer than its
1,799 distinct pixel images. These clusters explain why uncertainty cannot
be inferred from question count alone.

\begin{figure}[h]
  \centering
  \includegraphics[width=\textwidth]{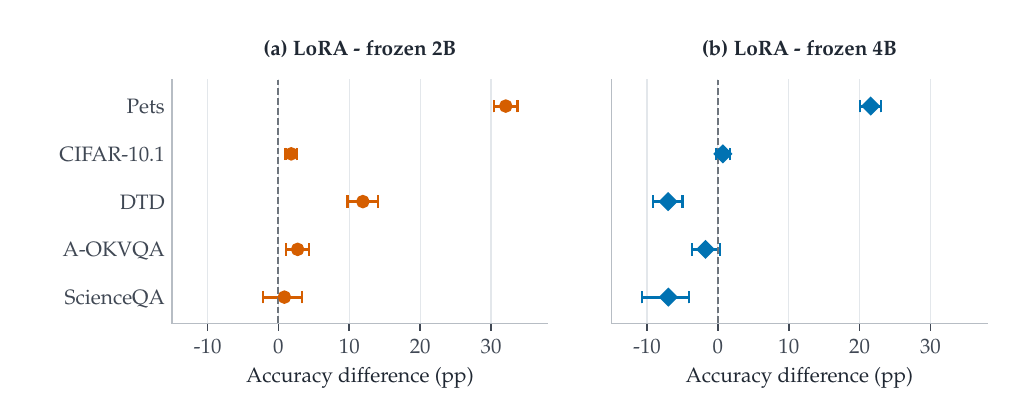}
  \caption{Adaptation is not a universal substitute for scale.
  Points are the source-selected seed-7 adapter's accuracy differences from
  (a) frozen 2B and (b) frozen 4B under the original user-only prompt.
  Whiskers are paired image-group bootstrap 95\% intervals, not
  optimization-seed SD, and are not multiplicity-adjusted.
  The same predesignated seed is used across tasks.}
  \label{fig:paired}
\end{figure}

\section{Instrument Checks and Readout Semantics}
\label{app:checks}
The adapted-interface diagnostic uses 74 predeclared Pets development
images and the source-selected seed-7 adapter.
The positive arm reaches 97.30\% accuracy.
Reversing the candidates preserves aggregate accuracy but changes some
decisions (97.30\% decision agreement).
The alternate instruction preserves every decision.
Substituting actual different-class images reduces accuracy to zero and
falls below the same predeclared 75\% criterion passed by the positive arm.
Direct and constrained one-token generation candidate logits match exactly
for every diagnostic example, with maximum absolute difference zero.

These checks establish functioning visual wiring and finite readout
equivalence, not arbitrary prompt invariance, universal image dependence,
or a measured speedup.

\section{Staged Research Milestones}
\label{app:agenda}
This appendix specifies proposed work, not completed experiments.
The order prioritizes gaps in the present claims over expanding benchmark
counts or model sizes.

\paragraph{Stage 1: extend the matched interface comparison.}
The prompt-only follow-up in Appendix~\ref{app:prompt-baseline} addresses
ordinary answer generation, direct candidate readout, and their measured
request costs for the fixed selected checkpoints.
Complete-candidate likelihood, broader prompt-selection budgets, and
training-objective controls remain open.
Compare candidate-conditioned fitting with ordinary
full-vocabulary answer SFT under matched supervision and update budgets.
Endpoints include accuracy, invalid outputs, log scoring, end-to-end latency,
and memory. If conventional VLM inference performs identically, the supported
benefit is interface convenience rather than a distinct algorithmic gain.

\paragraph{Stage 2: isolate schema and evidence sensitivity.}
Measure semantic-ID-aligned decisions and probabilities under candidate
permutations and equivalent descriptions. Vary candidate count and
distractor difficulty separately from the image distribution.
Adding candidates legitimately changes normalization, so numerical
probability invariance is not a valid universal requirement.
For visual dependence, compare relevant-region removal with equal-area
irrelevant-region controls alongside intact, mismatched, and noninformative
images. A grey-image gap alone cannot distinguish all distribution-change
effects from genuine use of question-relevant evidence.

\paragraph{Stage 3: test generalization across decision families.}
Extend to a small set of distinct families, such as spatial relations,
counting, and visual text, with entire families withheld from adapter
fitting. Split parent images and source groups before creating task variants.
Use language-only and candidate-only diagnostics for constructed options.
Match total supervision for joint-versus-task-specific adaptation and
repeat adaptation splits as well as optimization seeds. Report each
held-out family's outcome and source-task retention, not only a mixture
average. A new readout that supports more candidates must demonstrate
held-out cardinality behavior, not merely accept a larger tensor.

\paragraph{Stage 4: connect reliability to the action it serves.}
Define separate targets for answer correctness, availability of a valid
candidate, and sufficiency of the visual evidence.
Missing evidence may still permit a correct answer from context;
a valid-looking image can still elicit a wrong answer.
Fit rejection or re-observation thresholds only on separate calibration
data and test unchanged thresholds under shift.
Compare risk at matched coverage, or decision utility with stated error
and observation costs. A sufficiency detector earns its place through
better re-observation decisions, not through its detection AUROC alone.
Visual Jev's sufficiency study \citep{yu2026visualjev} makes this distinction
particularly important when designing the next experiment.

\paragraph{Stage 5: evaluate one bounded deployment workload.}
For example, choose among externally proposed UI objects and measure
proposal quality separately from selection quality.
Hold the downstream controller fixed when comparing task success,
unnecessary actions, latency, and memory.
For shared-image questions, compare shared-context serving with
already-batched and cached alternatives, including the design of
Visual Jev \citep{yu2026visualjev}. Shared execution is existing related
work, not a novelty claim for this roadmap.
Optimize the serving path only when workload measurements identify it
as the bottleneck; do not infer an advantage from the absence of generated
answer text.

\section{Matched Prompt-Only Baseline and Inference Cost}
\label{app:prompt-baseline}
\subsection{What the Factorial Does and Does Not Identify}
The two factors are checkpoint adaptation and inference/readout path:
frozen Qwen3.5-2B versus its existing classification adapter (source-selected
64-shot seed 7), crossed with unconstrained prompted generation versus
direct candidate logits. No new adapter or temperature is fitted.
This targeted follow-up retains the complete Pets and ScienceQA selections.
Pets tests a source task; ScienceQA tests question-conditioned transfer.
Readout conclusions from these two datasets are not generalized to all
seven original benchmarks.
The same native image processor, options in the same order, image budget,
task/question text, and format instruction apply to both paths.

This design estimates the effect of applying the already-trained adapter
under either readout and the effect of changing readout at fixed weights.
The interaction is
$(A_{\mathrm{LoRA,read}}-A_{\mathrm{LoRA,gen}})
 -(A_{\mathrm{frozen,read}}-A_{\mathrm{frozen,gen}})$.
Paired intervals resample the original image-similarity groups.
The design does not compare candidate-conditioned training with ordinary
full-vocabulary answer SFT at matched supervision and compute.
It cannot attribute all ``Jev-style'' effects to one algorithm or
extrapolate one selected adapter to training-seed uncertainty.

\subsection{Development-Only Format Selection}
The original principal prompt requests a letter in the user message.
A development-only comparison on eight preselected Pets images evaluated
two format cues: a system instruction and an assistant \texttt{Answer:} prefill.
Its rule preferred the system cue if every selected frozen response was
valid and used at most eight generated tokens, otherwise the prefill
if it met the same criterion. Both met the criterion; the predeclared
system-first rule was applied. Correctness did not select the prompt.
No evaluation-set generated response was inspected before this selection.

The fixed additional system instruction is:
\begin{quote}
You are a visual classifier. Choose one of the provided options.
Return exactly its case-sensitive letter, and nothing else.
\end{quote}
All four factorial cells are newly executed under this cue.
Consequently, these rows must not be spliced into the original user-only
results as though the prompts were unchanged.
Development examples do not enter the reported evaluation sets.

\subsection{Unconstrained Generation and Parsing}
Greedy generation uses one beam, caching, repetition penalty 1, thinking
disabled, and at most 128 new tokens. No candidate-vocabulary mask is
applied. It stops at either the native end-of-text or chat-end token,
or when a newline completes a valid
answer line. The fixed parser allows preceding explanation and recognizes
a case-sensitive letter on a complete line, optionally parentheses, a
final period, paired bold/backticks, or the prefixes \texttt{Answer:},
\texttt{Final answer:}, \texttt{The answer is}, and
\texttt{The correct answer is}.
Only one unique in-range answer is accepted; repeated identical answer
lines do not constitute conflicting choices.
Missing/conflicting answers, out-of-range letters, and cap truncation
are invalid and count as incorrect. There is no retry, label-name guessing,
or fallback to candidate logits.

Direct inference executes a separate forward pass with caching disabled,
rather than taking generation logits as its substitute.
First-generation-step logits are compared with direct inference for each
selected example.

All independently executed direct and first-generation-step candidate
logits match exactly in the final selected records.
Generation and direct decisions agree for both Pets checkpoints and frozen
ScienceQA. For adapted ScienceQA, the 55 disagreements are precisely the
55 invalid generated responses: 52 out-of-range letters and three
non-letter formats. There are no cap-truncated responses.
All valid generated choices match direct choices; the direct path answers
38 invalid cases correctly. These counts are based on every selected
example, not inferred from equal aggregate accuracies.
The parsing contract is intentionally fixed: the gap should not be
interpreted as the best achievable accuracy of a more permissive or
task-aware answer extractor.

\begin{table}[h]
  \caption{Checkpoint adaptation effects at fixed readout, and their
  interaction, in percentage points with paired image-group 95\% intervals.
  Intervals are descriptive and not multiplicity-adjusted.}
  \label{tab:prompt-effects}
  \centering
\begin{tabular}{@{}lccc@{}}
\toprule
Dataset & LoRA $-$ frozen: generation & LoRA $-$ frozen: readout & Interaction \\
\midrule
Pets & $+35.27\ [+33.47,+37.03]$ & $+35.27\ [+33.47,+37.03]$ & $+0.00\ [+0.00,+0.00]$ \\
ScienceQA & $+0.99\ [-1.18,+2.87]$ & $+2.88\ [+0.97,+4.68]$ & $+1.88\ [+1.13,+2.86]$ \\
\bottomrule
\end{tabular}

\end{table}

\subsection{Matched Warm Timing}
For each dataset, 64 images are selected by a deterministic,
hash-based ordering with seed 20260923, without consulting labels,
predictions, or times. Each path receives two warmups.
There are three measured repetitions per image; a deterministic random
permutation varies the order of the three paths within each image/repetition.
Batch size is one and the same checkpoint remains resident during its
comparisons. CUDA synchronization brackets each request.

The interval includes local image loading, decoding, pixel-hash validation,
prompt construction, preprocessing/tokenization, device transfer, inference,
and answer parsing or candidate-probability construction.
It excludes model loading, record writing, network, and serving queues.
Generation output-logit capture is disabled for timing.
Means first average repetitions per image; paired timing intervals resample
image-similarity groups. P95 is a descriptive percentile over warm requests.
The constrained-one-token control uses candidate-masked generation
limited to one token, with caching disabled.
It returns the same kind of bounded decision without ordinary
multi-token response termination.

Measurements use the pinned shared GB10 runtime, including its reference
PyTorch recurrent/convolution kernels. They are implementation-specific
warm request times, not exclusive-device service latencies or an optimized
generation-engine benchmark. The control distinguishes task-format and
termination costs from a supposed inherent advantage of ``not generating
one token.''

\begin{table}[h]
  \caption{Warm timing: each time cell is mean / P95 in milliseconds.
  Gen./direct and 1-token/direct are ratios of mean request times, not
  throughput improvements under batching. Every row uses the same fixed
  64 images and three repetitions per path.}
  \label{tab:prompt-timing}
  \centering
\begin{tabular}{@{}lrrrrr@{}}
\toprule
Dataset / weights & Direct (ms) & Generated (ms) & 1-token (ms) & Gen./direct & 1-token/direct \\
\midrule
Pets / Frozen & 127.9 / 147.2 & 170.0 / 188.8 & 131.7 / 149.3 & 1.33 & 1.03 \\
Pets / LoRA & 133.5 / 150.8 & 175.7 / 193.2 & 137.2 / 154.6 & 1.32 & 1.03 \\
ScienceQA / Frozen & 94.7 / 112.2 & 136.1 / 153.0 & 98.5 / 115.8 & 1.44 & 1.04 \\
ScienceQA / LoRA & 98.8 / 115.2 & 141.1 / 157.9 & 101.9 / 118.5 & 1.43 & 1.03 \\
\bottomrule
\end{tabular}

\end{table}

\begin{figure}[h]
  \centering
  \includegraphics[width=\textwidth]{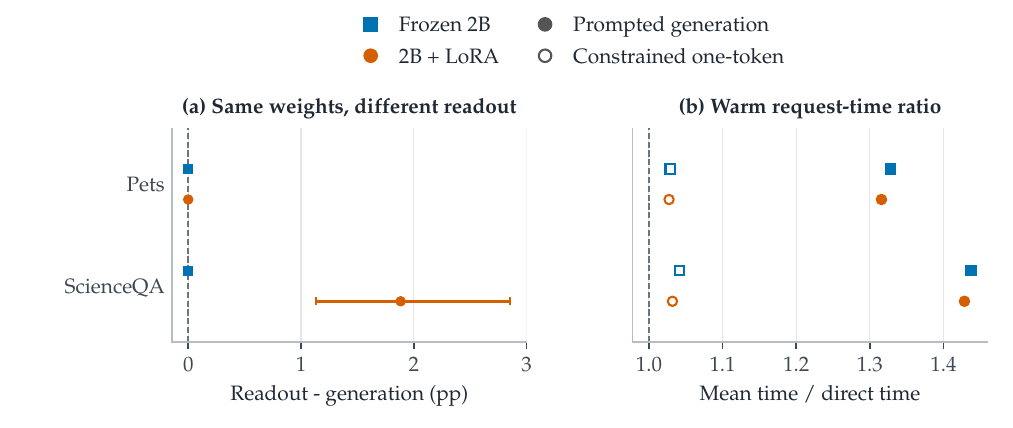}
  \caption{The inference trade-off at fixed weights.
  (a) Direct minus ordinary-generation accuracy, with paired image-group
  95\% intervals. (b) Ratios of mean warm times to direct inference;
  filled markers denote ordinary generation and open markers constrained
  one-token generation. Ratio markers are descriptive point estimates,
  not uncertainty intervals.}
  \label{fig:prompt-baseline}
\end{figure}
\end{document}